\documentclass[11pt,a4paper]{article}
\usepackage[hyperref]{acl2021}
\usepackage{times}
\usepackage{latexsym}

\usepackage{microtype}
\usepackage{graphicx}
\usepackage{subcaption}
\usepackage{enumitem}
\usepackage{booktabs,subcaption,amsmath}
\usepackage{multirow}
\usepackage{makecell}
\usepackage{tabularx}
\usepackage{soul}
\usepackage{hyperref}
\usepackage{cleveref}

\newcommand{\MainResultTable}{
\begin{table*}[h!]
\begin{center}
\small
\begin{tabular}[b]{p{5cm}cc}
\toprule
\bf Model & \bf CNN/DailyMail & \bf XSum \\
\midrule
BART-large~\cite{bart} & 44.16/21.28/40.90 & 45.14/22.27/37.25 \\
BERTSUM~\cite{liu-lapata-2019-text} & 43.85/20.34/39.90 & 38.81/16.50/31.27 \\
\midrule
\multicolumn{3}{l}{\textit{Previous evidence-based Extractive-Abstractive systems}} \\
\addlinespace
Bottom-Up~\cite{gehrmann-etal-2018-bottom} & 40.96/18.38/38.16 & - \\
SEAL~\cite{Zhao2020SEALSE} & 39.3/16.5/- & - \\
\midrule
\multicolumn{3}{l}{\textit{EASE (ours)}} \\
\addlinespace
Token-level sparsity 0.5 & 43.96/20.91/40.74 & 42.70/19.38/33.81 \\
Sentence-level sparsity 0.5 & 43.98/20.95/40.78 &  41.82/19.05/33.99\\
\bottomrule
\end{tabular}
\caption{ROUGE-1/2/L results for CNN/DailyMail and XSum.}
\label{tab:main_results}`
\end{center}
\end{table*}
}

\newcommand{\ModelSizeAblation}{
\begin{table}[t!]
\centering
\resizebox{\columnwidth}{!}{\begin{tabular}{c c c } \Xhline{2\arrayrulewidth}
            \textbf{Model} & \textbf{Token level} & \textbf{Sentence Level}  \\ \hline
              base Ex + base Ab & 42.9/19.8/39.7 & 42.42/19.7/39.27 \\
              base Ex + base Ab shared & 42.28/19.37/39.14 & 42.58/19.81/39.43 \\ \hline
              base Ex + large Ab & 43.9/20.8/40.7 & 43.88/20.83/40.69  \\ \hline
             \Xhline{2\arrayrulewidth}
\end{tabular}}
\caption{Ablation studies on the effect of model size and sharing. All models are trained with a sparsity of 0.5.}
\label{tab:model_size_ablation}
\end{table}
}

\newcommand{\ExtractorAblation}{
\begin{table}[t!]
\tiny
\centering
\resizebox{\columnwidth}{!}{\begin{tabular}{c c c} \Xhline{2\arrayrulewidth}
             \textbf{Extract Sentences} & \textbf{Sparsity 0.5} & \textbf{Sparsity 0.3} \\ \hline
              $\pi\%$ & 43.98/20.95/40.78 & 43.38/20.57/40.2  \\ 
              Top-3  & 41.37/18.58/38.18 & 41.19/18.63/38.01 \\
              Lead-3  & 40.84/18.16/37.71 & 40.6/18.05/37.46 \\
              Random-3  & 31.46/10.04/28.92 & 31.34/10.06/28.79 \\
             \Xhline{2\arrayrulewidth}
\end{tabular}}
\caption{Effect of different extraction techniques on the final summary.}
\label{tab:extractor_ablation}
\end{table}
}

\newcommand{\OverviewExample}{
\begin{figure}[t!]
\centering
\tiny
\begin{tabularx}{\columnwidth}{|X|}
\hline
\textbf{Source Document:} \\
\hl{\textbf{(CNN)Mike Rowe is coming to a river near you.}} "Sometimes, you hear about a person who makes you feel good about humanity, but bad about yourself," Rowe says. \hl{\textbf{On Thursday's episode of "Somebody's Gotta Do It," Rowe meets up with Chad Pregracke, the founder of Living Lands \& Waters, who does just that.}} Pregracke wants to clean up the nation's rivers one piece of detritus at a time. His quota? Always "more." Read Mike Rowe's Facebook post on how to break our litter habit. \hl{\textbf{Since he founded the nonprofit in 1998 at the ripe age of 23, Pregracke and more than 87,000 volunteers have collected 8.4 million pounds of trash from U.S. waterways.}} \hl{\textbf{Those efforts helped him earn the 2013 CNN Hero of the Year Award, along with numerous other honors.}} "Wherever you are, no matter if there's a stream, a creek, a lake, whatever, that needs to be cleaned up, you can do it. Just organize it and do it," he told CNN's Anderson Cooper after his win. Pregracke also gives Rowe a tour of the 150-foot, solar-powered barge that the Living Lands \& Waters staff calls home during lengthy cleanups. The part-home, part-office, part-dumpster has seven bedrooms, two bathrooms, a classroom and a kitchen -- and just happens to be made from a recycled strip club. According to the organization's latest annual report, Pregracke has made it his mission in 2015 to remove 500,000 more pounds of trash. If you'd like to help achieve this goal, visit his website to learn how to help: LivingLandsAndWaters.org/Get-Involved/.
 \\ \hline
\textbf{Summary:} Mike Rowe meets Chad Pregracke, the founder of Living Lands \& Waters. The nonprofit has collected 8.4 million pounds of trash from U.S. waterways. Pregracke was named the 2013 CNN Hero of the Year.
 \\
\hline
\end{tabularx}
\caption{An example of a summary and its explanation (highlighted) as generated by our framework.}
\label{tab:example}
\end{figure}
}

\newcommand{\TableHumanEvaluation}{
\begin{table}[t!]
\centering
\resizebox{\columnwidth}{!}{\begin{tabular}{c | c c | c} \Xhline{2\arrayrulewidth}
            \textbf{Models} &  \multicolumn{2}{c|}{Summary} & Extraction\\
              &  Consistency & Relevance & Relevance \\
          \Xhline{2\arrayrulewidth} 
          BART   & \textbf{4.89} & \bf{4.13}  & - \\
          Token-level model   &  4.77 & \textbf{4.16} & - \\
          Sentence-level model   &   \bf{4.86} & 3.80 & \textbf{4.45} \\
          Lead-3 extraction & - & - & 4.38 \\
            \Xhline{2\arrayrulewidth}
\end{tabular}}
\caption{Human Evaluation results on CNN/DM. We evaluate our token-level and sentence-level models, with a sparsity of $0.5$ on summary relevance and consistency and compare with BART. We evaluate extraction relevance of our sentence-level model and compare with Lead-3.}
\label{tab:table_human_evaluation}
\end{table}
}

\newcommand{\PretrainigResults}{
\begin{table}[t!]
\centering
\resizebox{\columnwidth}{!}{\begin{tabular}{c c c } \Xhline{2\arrayrulewidth}
            \textbf{Model} & \textbf{Token level} & \textbf{Span Level (lasso)}  \\ \hline
               vanilla EASE &  43.96/20.91/40.74 & 44.33/20.67/41.06 \\
              + pretraining  & 44.12/20.89/40.80 & 44.06/20.82/40.83  \\ \hline
    
             \Xhline{2\arrayrulewidth}
\end{tabular}}
\caption{CNN/DM results on token-level models trained with lasso loss and pretraining.}
\label{tab:pretraining}
\end{table}
}

\newcommand{\SSLResults}{
\begin{table}[t!]
\centering
\resizebox{\columnwidth}{!}{\begin{tabular}{c c c } \Xhline{2\arrayrulewidth}
            \textbf{Model} & \textbf{Token level} & \textbf{Sentence level}  \\ \hline
               vanilla EASE &  43.96/20.91/40.74 & 43.98/20.95/40.78  \\
              + SSL  & 44.28/21.21/41.0 & 44.10/21.12/40.89 \\ \hline
    
             \Xhline{2\arrayrulewidth}
\end{tabular}}
\caption{Results on token level and sentence level models, trained with additional semi-supervised extraction.}
\label{tab:ssl}
\end{table}
}

\newcommand{\addExamples}{
\begin{figure*}

\tiny
\centering
\begin{tabularx}{\linewidth}{|X|}
\hline
\textbf{Source Document:} \\
\hl{\textbf{(CNN)Two passengers found dead on a cruise ship in Puerto Rico appear to have died in a murder-suicide, the cruise line said.}} \hl{\textbf{Holland America Line said two guests were found dead inside their stateroom on the ms Ryndam at 11:30 a.m. Thursday.}} "The cabin was immediately secured, and the authorities were notified, including the FBI," Holland America said. "We are cooperating fully with the investigation, and the authorities will make the official determination on what occurred." \hl{\textbf{FBI spokesman Moises Quinones said authorities were on scene investigating.}} The ship left Tampa, Florida, on March 29 on a 14-day Southern Caribbean cruise. It's currently in San Juan, Puerto Rico. Puerto Rico Port Authority spokesman Efraín Santiago told El Nuevo Dia newspaper that the cleaning staff on the ship had discovered the deceased passengers after knocking on the cabin's door.
\\ 
\textbf{Summary (Sparsity 0.3):} Holland America Line said two guests were found dead inside their stateroom on the ms Ryndam at 11:30 a.m. Thursday. The FBI is investigating. \\
\hline
\textbf{Source Document:} \\
\hl{\textbf{(CNN)Gastrointestinal illness has gripped 100 people on the cruise ship Celebrity Infinity, according to a report from the Centers for Disease Control.}} \hl{\textbf{Of the ship's 2,117 passengers, 95 have suffered from vomiting, diarrhea and other symptoms, the CDC said.}} \hl{\textbf{The illness has also affected five members of the 964-person crew.}} \hl{\textbf{The CDC has yet to determine what's causing the ailments.}} Two staffers from the agency are scheduled to meet the West Coast-based ship in San Diego on Monday. The Infinity left San Diego on March 29. It made its last stop in Puerto Vallarta, Mexico, on April 10, according to MarineTraffic.com. \hl{\textbf{Celebrity Cruises has been taking action since the outbreak began, including increasing cleaning and disinfection procedures, keeping passengers informed and taking specimens from the afflicted for testing by the CDC, the agency says.}} According to the Maritime Executive, this is the third time the Celebrity Infinity has suffered an outbreak of gastrointestinal illness, with others occurring in 2006 and 2013. The ship was built in 2001 and refurbished in 2011.
\\ 
\textbf{Summary (Sparsity 0.5):} Of the ship's 2,117 passengers, 95 have suffered from vomiting, diarrhea. The illness has also affected five members of the 964-person crew. Celebrity Cruises has been taking action since the outbreak began.
 \\
\hline
\end{tabularx}
\caption{Summarization outputs, along with their explanations (highlighted), from our systems at different sparsity levels.}
\label{fig:summary_output}
\end{figure*}
}

\aclfinalcopy % Uncomment this line for the final submission
\newcommand\blankfootnote[1]{%
  \let\thefootnote\relax\footnotetext{#1}%
  \let\thefootnote\svthefootnote%
}
\title{EASE: Extractive-Abstractive Summarization with Explanations}
\author{
  Haoran Li \thanks{\ \ Equal contribution.} \quad Arash Einolghozati\footnotemark[1] \quad Srinivasan Iyer \\
  \textbf{Bhargavi Paranjape \quad Yashar Mehdad \quad Sonal Gupta \quad Marjan Ghazvininejad} \\
  Facebook \\
  \texttt{\{aimeeli,arashe,sviyer\}@fb.com} \\
  \texttt{\{bparan,mehdad,sonalgupta,ghazvini\}@fb.com} \\
}
\date{}

\begin{document}
\maketitle
\begin{abstract}
%they do not provide an explanation for the generated summaries,
Current abstractive summarization systems outperform their extractive counterparts, but their widespread adoption is inhibited by the inherent lack of interpretability. To achieve the best of both worlds, we propose EASE, an extractive-abstractive framework for evidence-based text generation and apply it to document summarization. We present an explainable summarization system based on the Information Bottleneck principle that is jointly trained for extraction and abstraction in an end-to-end fashion. Inspired by previous research that humans use a two-stage framework to summarize long documents~\cite{decomposition_human_summary}, our framework first extracts a pre-defined amount of evidence spans as explanations and then generates a summary using only the evidence. Using automatic and human evaluations, we show that explanations from our framework are more relevant than simple baselines, without substantially sacrificing the quality of the generated summary. 

\end{abstract}

\section{Introduction}
Pretrained sequence-to-sequence language models such as BART~\cite{lewis-etal-2020-bart}, T5~\cite{t5} and their variants have achieved state-of-the-art results on various tasks such as summarization, machine translation, and data2text tasks~\cite{pegasus, kale-rastogi-2020-text}. Despite the higher fidelity compared with models without pretraining for tasks such as summarization~\cite{maynez-etal-2020-faithfulness}, the lack of explainability in such generation models remains an obstacle to their broader adoption. 
\OverviewExample
Extractive summarization systems, on the other hand, have the advantage of being interpretable but are too restrictive by forcing the output to be spans from the document, reducing their naturalness and conciseness. In this paper, we propose EASE, a novel framework that combines the two systems to produce natural and interpretable summaries. Our other motivation is for the generated summary to reuse words from the source document whenever possible to make it less prone to misrepresentation. To that end, we primarily study extractive-like datasets such as CNN/DailyMail. 

The existing explainable extractive-abstractive systems can be divided into three main categories: 1- Relying on attention for explainability~\cite{hsu-etal-2018-unified}. Due to the probabilistic nature of the attention mechanism, it falls short of providing usable evidence; 2- Providing word-level evidence for the generated summaries~\cite{gehrmann-etal-2018-bottom}; and 3- Training the content selector separately using pseudo labels or other heuristics~\cite{liu-lapata-2019-text, pilault-etal-2020-extractive}. Though more useful than attention, this evidence is too granular to be useful for humans. In contrast, we seek a theoretically-grounded model that can learn the evidence extraction end-to-end.

Perhaps the closest work to ours is \citet{Zhao2020SEALSE} focusing on long-document summarization by training a joint extractive-abstractive model via weak supervision, though they report poor results on benchmarks such as CNN/DM.
EASE on the other hand, is based on the Information Bottleneck (IB) principle~\cite{IB}, which formalizes the trade-off between the size of the extracted evidence and the information provided for the generation of the final output. While this method has been successfully adopted by prior work for rationale extraction in discriminative tasks~\cite{paranjape-etal-2020-information}, we extend it to generative tasks where the extracted evidence can be viewed as a coarse version of the final abstractive output. 

We leverage pretrained language models that first extract the necessary evidence from the source document (\textit{extractor}) and then, using only the extracted evidence spans, generate the final output (\textit{abstractor}). 
While we focus on the summarization task to evaluate EASE, it is generic and can be adopted for other generative tasks such as evidence-based QA. Fig.~\ref{tab:example} shows an example of the explanations and summary generated by our system.

Our main contributions are as follows: 
\begin{itemize}
\item We propose EASE, a general-purpose theoretically-grounded Extractive-Abstractive framework for explainable text generation that is jointly trained for extraction and abstraction in an end-to-end fashion. We apply EASE to text summarization.
\item The abstractor generates the summary using only the extracted evidence spans; the evidence spans are `explanations' for the generated summary. We propose a new \textit{sparsity budget} parameter that controls the length of the explanations/evidence spans.
\item Our results show that EASE extracts explanations better than the baselines without significantly sacrificing the quality of the generated summary, compared with the state-of-the-art fully abstractive systems on the CNN/DailyMail dataset.

\end{itemize}

We show that our sentence-level extractive-abstractive system performs on-par with BART in an automatic evaluation on CNN/DailyMail while also generating explanations. In human evaluation, EASE is similar to BART in terms of summary \textit{consistency} but slightly lower in terms of \textit{relevance}, which is not very surprising because our framework restricts the information to the abstractor by limiting its input to only the evidence spans. We believe that the interpretability provided by the explanations outweighs the slightly lower quality, as it makes summarization systems more practical (e.g., by clicking-through to the source), and thus, more viable for wider adoption.

\section{Extractive-Abstractive Framework}

Some evidence exists that humans use a two-stage extractive-abstractive framework to summarize long documents~\cite{decomposition_human_summary} by first extracting salient parts and then deciding what to eliminate, reword, and reorganize.
We propose EASE, a framework that learns extraction and abstraction collectively in an end-to-end fashion. This not only provides an explanation for the generated summary, which can be many times smaller than the original document, but also reduces the effective input length used during abstraction. This has been shown to directly correlate with the extent of hallucination in pretrained language  models~\cite{bart_webnlg}.

%We hope that our pretraining, which can be looked at as summarization with compression rate of 0,  can help the downstream summarizer in both of these tasks. Thus pre-tranining includes: 1- Extract the important information 2- Reconstruct the original document

In order to formalize the problem, we use the IB principle to learn an optimal model between the original document $x$ and the final summary $y$ through a compressed representation $z$. 
The IB objective is to minimize the following:
\begin{equation}
    L_{IB}= I(x;z) - \beta I(z;y),
    \label{eq:ib}
\end{equation}
where $I()$ is the mutual information. This objective encourages $z$ to contain only the information about $x$ that is useful in predicting $y$. Moreover, $\beta$ controls the trade-off in $z$ between containing information about $x$ (i.e., sparsity) vs about $y$ (i.e., prediction quality).

\begin{figure}[t]
\includegraphics[width=\columnwidth]{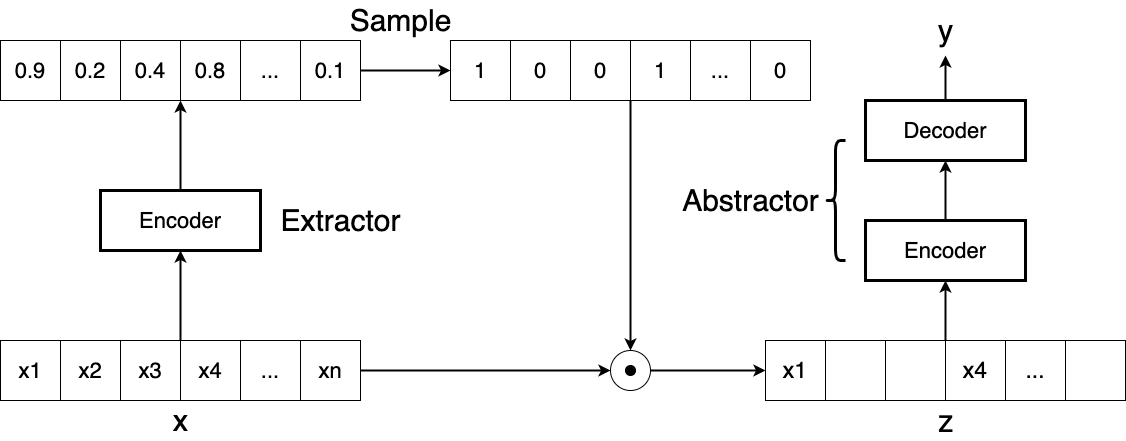}
\caption{The Extractive-Abstractive model architecture. The extractor samples the evidence from the source which is used by the abstractor.}
\label{fig:ex_abs}
\end{figure}

 We use a relaxation for~(\ref{eq:ib}) similar to~\citet{paranjape-etal-2020-information} to make it tractable. As such, $z$ is obtained by masking the original document $x$ to produce a summaries $y$. We illustrate EASE in Fig.~\ref{fig:ex_abs}. EASE can perform extraction (i.e., masking) either at the token or at the sentence level. We first describe the token-level model and subsequently generalize it for sentence-level extraction. As such, the extractor masks tokens in the original document $x$ to extract a rough summary $z$, which is used as evidence by the abstractor to produce the summary $y$.  We define $z=m \odot x$ where $m$ is a boolean mask on the input $x$. This is similar to the masking process used in Masked Language Models (MLM), except that instead of random masking~\cite{devlin-etal-2019-bert} or heuristic-based masking~\cite{pegasus, zhang-etal-2019-ernie}, we learn which tokens should be masked in an end-to-end fashion. Using the variational bound~\cite{alemi_variational} on~(\ref{eq:ib}), the model is trained using two loss terms. The first loss ensures that the final summary is close to the golden summaries:
\begin{equation}
\label{eq:task_loss}
L_{Task}=E_{m\simeq p(m|x)} [-\log q_{\theta}(y|m \odot x)],
\end{equation}
where $q_{\theta}(y|z)$ is a parametric approximation to the true likelihood $p(y|z)$.

Similar to \citet{paranjape-etal-2020-information}, we assume that the mask variables over individual words are conditionally independent given the input $x$. This means that the evidence $z$ can contain redundancies, as the extractor chooses evidence individually without conditioning on prior extractions. Since the extracted evidence is not the final summary, the abstractor still has the opportunity to eliminate redundancies. \citet{Summarunner} explore a modeling approach that keeps track of the current state of the summary, but we leave this direction to future work. Formally,

\begin{equation*}
    p_{\theta}(z|x) = \prod_{j}  p_{\theta}(z_j|x),
\end{equation*}
where $p_{\theta}(z_j|x)=Bernoulli(\theta_j(x)).$

Optimizing the loss in~(\ref{eq:task_loss}) would result in the extractor masking no tokens and hence, maximizing the mutual information between the input and output of the abstractor. Therefore, the second loss term is a sparsity constraint to ensure that the extractor's output is a measurable subset of input tokens and can be used as evidence for the abstractor output:
\begin{equation}
\label{eq:sparsity}
L_{Sparsity} = \sum_{j} KL[p_{\theta}(z_j|x),r(z_j)],
\end{equation}
where we set the prior distribution $r(z_j)=Bernouli(\pi)$. For summarization tasks $\pi$ can be small i.e. $0.3\leq \pi \leq 0.5$.
%\citet{paranjape-etal-2020-information} showed that the KL divergence with a sparse prior yields more accurate sparsity in practice than using a norm loss on the number of unmasked tokens~\cite{bastings-etal-2019-interpretable}.
As such, the combined loss can be written as:

\begin{equation}
\label{eq:loss}
\begin{split}
    L_{EA}= & E_{m\simeq p(z|x)} [-\log q_{\theta}(y|m \odot x)] \\
    + &\beta \sum_{j} KL[p_{\theta}(z_j|x),Bernouli(\pi)],
\end{split}
\end{equation}
where  $p_{\theta}(z|x)$ is the parametric posterior distribution over $z$ and $\beta$ is a hyperparameter to weigh the performance-sparsity trade-off. 
% In practice, we found out that the performance was robust for values of $beta$ between $1$ and $10$. We picked $\beta=5$ for our experiments.

\subsection{Soft Masking}

The combined loss presented above is not differentiable, as it includes sampling operations from Bernoulli distributions. Since we aim to learn the masking function (unlike random masking), this would not be amenable to end-to-end training using backpropagation. Rather than using the REINFORCE algorithm which suffers from high variance \cite{bastings-etal-2019-interpretable}, we use the Gumbel Softmax reparameterization trick~\cite{Gumbel_softmax} similar to~\citet{paranjape-etal-2020-information}. 
This replaces the sampling step with an argmax: $argmax_{i\in {0,1}}(log p(z_j|x) + g_i)$, where $g_i$ is a random sample from the Gumbel$(0,1)$ distribution. Finally, the argmax is replaced by a weighted softmax:

$$z^*_j= \frac{\exp{((\log(p(z_j=1|x)+g_1)/\tau)}}{\sum_{i \in {0,1 }}\exp{((\log(p(z_j=i|x)+g_i)/\tau)} }.$$
Note that $z^*_j \in (0,1)$  gets boundary values (i.e., 0 or 1) when $\tau\rightarrow 0$ (in practice, we use $\tau=0.01$).

% \begin{figure}[t]
% \includegraphics[width=\columnwidth]{plots/sampling.png}
% \caption{}
% \end{figure} 

\subsection{Model Architecture}
As illustrated in Fig.~\ref{fig:ex_abs}, our model has two parts: the extractor and the abstractor. 
The extractor is a pretrained transformer encoder similar to BERT~\cite{devlin-etal-2019-bert} with an additional linear layer on top that computes $p_\theta(z_j|x)$.
The abstractor on the other hand, is a pretrained seq-to-seq language model like BART~\cite{lewis-etal-2020-bart}. From our experiments, we find a BART-base encoder (6 layers) to be adequate as an extractor model, while we use a BART-large abstractor. Note that after evidence extraction, to ensure that there is no leakage of information, we need to encode the extracted tokens separately again. Using the same encoded representation would leak information to the abstractor about the masked tokens.

During training, given an input $x$, the extractor generates a probability for each token in $x$ to be selected (i.e. not masked). Based on these probabilities ($p_j$), we soft-sample $m_j$ with values in (0,1). We then pass $z=x\odot m$ to the abstractor to generate the output. In our experiments, we tried two different ways of masking the input using $m$: 1) directly masking the embedding, i.e. $z_j = m_j*x_j+(1-m_j)*x_{mask}$ where $x_{mask}$ is initialized from the BART's original \texttt{<mask>} token, and, 2) using $m$ as an attention mask for both the encoder's self attention as well as the encoder-decoder cross attention, i.e. to block attention to the masked tokens. However, we did not observe a significant difference between these two schemes.
During the inference, the extractor deterministically selects the top $\pi\%$ of the source tokens. Such hard masking ensures that the sparsity requirement is exactly met during inference time.

% \subsection{Implementation Details}

% We experimented with sharing the parameters of the abstractor's and extractor's encoders but found out that the task loss's effect on the extractor (which should pass through the Gumbel softmax) is much less than the task loss on the abstractor's encoder. As such, the extractor is poorly trained to pick the most informative pieces of text. 

% We experimented with both N scaling the embedding vectors as well as attention masks similar to~\cite{paranjape-etal-2020-information} in which we mask the attention heads from the masked tokens. We did not observe a significant difference between these two schemes.

\subsection{Sentence-level Extraction}
In the previous section, we described token-level extraction where each token in the source document is individually masked or retained. The main drawback of using scattered token-level extraction is that it is difficult to be used as interpretable evidence. While in Section \ref{sec:lasso}, we explore a method for improving the explainability of token-level evidence by encouraging span-level extraction, in this section, we focus on sentence-level extraction as an effective means to achieve explainability.

In sentence-level extraction approaches, the model first selects the sentences that need to be masked, followed by the masking of all tokens within those sentences. Unlike the token-level model, the extractor's output in this setup is a linguistically plausible (but possibly redundant) extractive summary, i.e., complete sentences from the source. 
For sentence-level extraction, we add a special \texttt{[CLS]} token to the beginning of each sentence and use its representation as the sentence encoding. We also add a segment embedding to each token in the sentence to distinguish between the sentences in a document. The segment embeddings are initialized randomly and learned during training. We use the \texttt{[CLS]} token representation to perform soft masking as in the token-level model. 
% The mask is then applied to all tokens in the sentence.  

\section{Experimental Settings}
\paragraph{Datasets:}
We primarily experiment with the CNN/DailyMail dataset~\cite{CNN_2015} owing to its extractive-like nature; its summaries are typically closely related to the source sentences. We also present results on the XSUM~\cite{narayan-etal-2018-dont} dataset, a highly abstractive dataset in which summaries can be viewed as a title for the source documents. 

\paragraph{Model Hyperparameters and evaluation metrics:}
We initialize the seq-to-seq abstractor with the BART~\cite{lewis-etal-2020-bart} model (base or large), and initialize the extractor using the BART-base encoder.
\par \noindent
We use the fairseq codebase\footnote{https://github.com/pytorch/fairseq} for our experiments and use the same hyperparameters as used for fine-tuning BART on CNN/DM and XSum by the official codebase. Specifically, we fine-tune BART using a polynomial decay learning rate scheduler with the Adam optimizer \citep{adam}. We use a learning rate of 3e-5 with 500 warmup steps and train for 20000 steps. During our initial experiments, we observed similar results for values of $\beta \in [1,10]$ in~(\ref{eq:loss}). We use $\beta=5$ in our reported results. We use ROUGE F1 scores (R1/R2/RL) for the automatic evaluation. ROUGE scores were calculated using the files2rouge toolkit\footnote{https://github.com/pltrdy/files2rouge}.

\section{Results}
\MainResultTable
In this section, we report the performance of our model from both automatic and human evaluation perspective, along with ablation studies. Figure~\ref{fig:summary_output} shows example summaries along with their explanations from our system at different sparsity levels.
\subsection{Automatic Evaluation}
In Table~\ref{tab:main_results}, we present the performance of our model for CNN/DM and XSum when using a sparsity of $0.5$, with a BART-base encoder as the extractor and a BART-large abstractor. We also present the performance of BART and BERTSUM as representative abstractive and extractive systems, respectively. Moreover, they can be considered as EASE's exctractor (BERTSUM) or abstractor (BART) on their own.  Note that for BERTSUM, we present the performance of the Ext-large version for CNN/DM and the two-stage ExtAbs version for XSum. We also include results from previous evidence-based extractive-abstractive systems for comparison. For CNN/DM, our token-level and sentence-level models that use around $50\%$ of the source input perform slightly better than BERTSUM, but slightly worse than BART-large. 
For XSum, our gap with the BART-large baseline is larger. This is expected given that XSum summaries are highly abstractive, making it much harder for the extractor to extract the most important information in an end-to-end fashion.  

Moreover, we observe that the sentence-level model performs slightly better than the token-level model for CNN/DM but slightly worse for XSum. We hypothesize that for the more extractive CNN/DM dataset, keeping continuous spans of text is of paramount importance, while for the more abstractive XSum dataset, the sparsity budget can be better spent on a more scattered extraction of key pieces throughout the document.
In section \ref{sec:further_improvement}, we explore ideas to 1) improve the performance of the token-level model using pre-training; 2) improve the explainability of token-level models by encouraging the extraction of continuous spans; and 3) improve the performance of both token and sentence level models using semi-supervised learning.

\subsection{Model Analysis}
\paragraph{Effect of Sparsity Prior:}
\begin{figure}[t]
\centering
\includegraphics[width=\columnwidth]{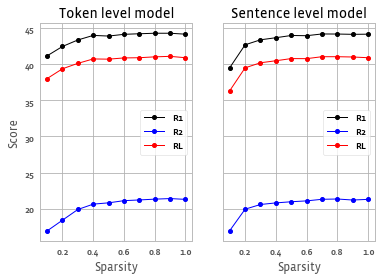}
\caption{R1/R2/RL vs Sparsity for token level and sentence level models. For sentence level model, we enforce it to extract at least three sentences.}
\label{fig:sparsity}
\end{figure}
In this section, we investigate the effect of sparsity on the generated summaries. Figure  \ref{fig:sparsity} presents ROUGE score of both token-level and sentence-level models, trained with different sparsity priors. As expected, increasing the sparsity ratio improves the ROUGE scores at the cost of more verbose explanations. Moreover, the performance gains flatten after a sparsity of around $0.3$. We found that token-level models are more robust to lower sparsity rates, i.e. they can remove functional words without losing document information, but they are not well-suited in terms of explainability. Note that for the sentence-level models, at inference time we extracted at least three sentences to ensure that short documents would have enough evidence at lower sparsity rates. 
% \par \noindent
\paragraph{Effect of model size:}
We examine the effect of using models of different sizes on summarization performance, and also explore the possibility of sharing the encoder. We consider BART-base and BART-large for the abstractor. We also experimented with using the BART-large encoder for the extractor but found it very unstable and hard to tune the relative loss weights. To explore the possibility of reducing the model size, we also experiment with sharing the encoder's parameters between the extractor and abstractor encoders. Table \ref{tab:model_size_ablation} presents results of these settings for both token-level and sentence-level models using a sparsity of $0.5$. 

%  On the other hand, Using a large extractor yields poor performance on sentence level models. We find that the information loss has a particularly large effect on the sentence representation (representation of the corresponding BOS tokens through transformer encoder). Since the information loss is propagated through these BOS tokens, while the task loss's effect on the extractor is felt, after it propagates through the entire abstractor and the Gumbel Softmax. Thus, using a large extractor with more layers is unstable and difficult to tune.
 
We can see that using a large model for the abstractor yields significant improvements. Moreover, sharing the encoder between the extractor and the abstractor does not hurt the performance. However, since using a large abstractor is essential while using a large extractor is unstable during training, we use a BART-base extractor and a BART-large abstractor for our default setting.
\ModelSizeAblation
% \par \noindent
\paragraph{Effect of extraction:} We evaluate the effect of extraction quality on the final summary for our sentence-level models. We use our model trained with different sparsity rates but during inference, feed only the top-3 sentences with highest scores to the abstractor for generating the summary. We compare with the baselines of using random-3 and  lead-3 sentences as well as using all $\pi\%$ of sentences. Table \ref{tab:extractor_ablation} presents results of our two models with sparsity values of $0.5$ and $0.3$. We find that for both models, summaries using the top-3 sentences selected by the extractor outperform lead-3 extraction, even though the CNN/DM dataset has a strong lead bias. We conclude that our extractor is indeed extracting important sentences, which we further confirm using human evaluations, described in the next section.
\ExtractorAblation

\subsection{Human Evaluation}
We conduct human evaluation on both the extracted evidence and the generated summaries. For the summaries, we asked annotators to rate them between 1-5 on two qualitative aspects of the summary: Consistency and Relevance. Consistency is the factual alignment between the summary and the source document, measuring whether the summary is changing details or hallucinating. Relevance measures whether the summary captures the key points of the source document. We compared our generated summaries with BART as a baseline. We also evaluate the relevance of extractions from the sentence-level models. To make evaluation easier, we gather the top-3 sentences with the highest extraction scores and ask annotators whether those are the most important sentences in the source document. Here, we compare with Lead-3 extraction as a baseline. 

We sampled 200 examples from the CNN/DM test set and conducted human evaluation using Amazon Mechanical Turk with three annotators. We present the average annotators' scores in Table~\ref{tab:table_human_evaluation}, using z-score p-values smaller than $0.01$ to measure statistical significance. We find that for extraction relevance, the top-3 sentences from our extractor scored higher than Lead-3, which itself received a high relevance score due to the strong lead bias in the CNN/DM dataset. For abstractive summaries, we find that the sentence-level model achieves a similar consistency score as BART, but slightly better than the token-level model. On one hand, the sentence model achieves a lower relevance score than BART and token model. We hypothesize that the explainable nature of the sentence model results in a loss of some of the key information in the source document as expected, whereas the token model avoids this by extracting keywords throughout the source. On the other hand, the token-level model can fabricate new details between the extracted keywords, which results in lower consistency. As such, there is an inherent trade-off between relevance and explainability.

\TableHumanEvaluation
%\subsection{Human Evaluation}
\addExamples
\section{Further improvements and Future Work} \label{sec:further_improvement}
\subsection{Span-level model with Lasso loss} \label{sec:lasso}
In the previous section, we found that although sentence-level models are explainable, they can miss out on key parts of the source document. However, token-level models enjoy much more freedom during extraction but yield evidence that is not very useful for humans. To find a compromise between these two, i.e. a span-level model, we attempt to make the evidence extracted by token-level models more contiguous, by adding a lasso loss~\cite{bastings-etal-2019-interpretable} to the total loss in~(\ref{eq:loss}):
\begin{equation}
 L_{Lasso} = \sum_{i=0}^{n-1} |z_i-z_{i+1}|,
\end{equation}
where $n$ is the number of source tokens. The lasso loss ensures that the number of transitions between the masked and unmasked tokens is minimized and hence, the model extracts more contiguous spans of text as evidence. In the first row of Table~\ref{tab:pretraining}, we observe that the lasso loss mainly improves the token-level model. This is particularly evident in the improvement in $R_L$  which is due to the extraction of contiguous spans as evidence.
% As such, the overall loss for the token-level model can be written as
% \begin{equation*}
% \label{eq:loss_token}
% \begin{split}
%     L_{EA}= & E_{m\simeq p(z|x)} [-\log q{\theta}(y|z \odot x)] \\
%     + &\beta_1 \sum_{j} KL[p_{\theta}(z_j|x),Bernouli(\pi)] \\
%     + & \beta_2 \sum_{i=0}^{n-1} |z_i-z_{i+1}|
% \end{split}
% \end{equation*}
% where $\beta_1$ and $\beta_2$ are hyperparameters for our experiments

\subsection{Unlabeled Pretraining}

Although we initialize the extractor and abstractor with pretrained language models, the model may benefit from further pretraining suited to the downstream task.
To this end, we use our model in an auto-encoding fashion, i.e., the abstractor reconstructs the original text using the extracted pieces selected by the extractor. Our hypothesis is that an extractor capable of extracting the most informative parts from which the source can be reconstructed should be better positioned to extract important parts of the source, resulting in higher-quality summaries. 
Therefore, we pretrain EASE on the WikiText-103~\cite{wikitext} dataset to reconstruct the original unlabeled documents using the same loss as in~(\ref{eq:loss}) by setting $Y=X$. This can be viewed as a special case of summarization, where the compression rate is one. We only pretrain the token-level model, since pretraining sentence-level models without measures such as topic guidance~\cite{kang-hovy-2020-plan} typically leads to hallucination.
Results on the CNN/DM dataset by adding pretraining are presented in the second row of Table~\ref{tab:pretraining}. Even though pretraining improves the token-level model, results for the span-level model are mixed. Our hypothesis is that the lasso continuity helps with summarization by picking contiguous spans, as evidenced by the high $R_L$. However, during the reconstruction pretraining, the lasso loss can be problematic by masking long spans, which are then prone to causing hallucinations. We leave pretraining alongside span extraction using techniques such as guided reconstruction to future work.

\PretrainigResults

\SSLResults 

\subsection{Semi-supervised Training}

Multiple recent works \cite{Summarunner,liu-lapata-2019-text} have explored heuristics to obtain pseudo alignments between target summaries and source sentences for summarization datasets. To evaluate the effect of weakly supervising the extractor in EASE using these pseudo labels, we use the greedy procedure of \newcite{liu-lapata-2019-text} to obtain oracle extractive annotations for CNN/DM. As such, we maintain an evidence set and greedily add source sentences to the set that yield the maximum increase in its ROUGE score against the target summary. This yields a binary labeling of input sentences and we introduce an additional binary cross entropy loss to our training objective in~(\ref{eq:loss}) between this binary labeling and the predicted masking probabilities.  By using the sentence-level pseudo labels for the tokens of each sentence, we also add this loss to the token-level models. We have shown the results in Table.~\ref{tab:ssl}. We observe improvements in all ROUGE metrics for both sentence-level and token-level models, though the gains on the former are more modest. Studying the interaction of this objective with the aforementioned lasso objectives is left for future work. 

%We examine the effect of directly providing the extractor with pseudo evidence labels. 

\section{Related Work}

\subsection{Pretrained Models for Summarization}

% \citet{Khandelwal_pretrained} were among the first to propose a simple LM objective to pretrain a transformer for summarization.
\citet{lewis-etal-2020-bart} introduced BART, a general-purpose denoising seq2seq transformer, that achieved the state-of-the-art results on many summarization tasks. Later, \citet{pegasus} extended the MLM denoising objective using sentence masking. 
% \citet{Goodman2019MultistagePF} adds the copy mechanism and content selection on top of a pretrained language model and sequentially train on various datasets. 
\citet{zhang-etal-2019-hibert} introduced a multi-stage encoder for extractive summarization, whereas \citet{zhang-etal-2019-pretraining} use a two-stage decoder to generate summaries by  creating a draft and refining it using a pretrained language model.
In EASE, we use pretrained models, i.e., BART, to initialize the extractive and abstractive modules but after that, use an end-to-end loss that trains both modules simultaneously.

\subsubsection{Self-supervised Summarization}

\citet{miao-blunsom-2016-language} introduced an autoencoder setup for sentence compression to reduce the need for labeled examples. A copy ptr/generator model was used for the compressor which alongside the reconstructor is trained to reconstruct the unlabeled documents. Moreover, REINFORCE~\cite{williams_reinforce} was used to train the model end-to-end. 
~\citet{baziotis-etal-2019-seq} introduced a similar autoencoder setup but used the Gumbel Softmax reparametrization for training. \cite{fevry-phang-2018-unsupervised} also used a denoising autoencoder to compress sentences and a countdown at the decoder to control summary length.

Inspired by the IB principle,~\citet{west-etal-2019-bottlesum} introduced a recursive algorithm to prune a document to form an unsupervised extractive summary.  These summaries are in turn used to train a self-supervised system using a next-sentence objective is used. In contrast, we use a loss formulation derived directly from the IB  and train the model end-to-end.
\cite{saito_saliency} used a saliency model to extract important pieces of a document before feeding them to an abstractive seq2seq model. In contrast with our model, the saliency module is trained separately by using heuristics to provide pseudo labels for the extraction.
\cite{yang-etal-2020-ted} proposed pretraining over millions of news articles using the lead sentence as the self supervision.
% \cite{wang-etal-2019-self} introduced auxiliary pretraining tasks such as sentence mask prediction to capture the document-level context in extractive summarization.

\subsection{Extractive-Abstractive Summarization}
The transformer decoder~\cite{generating_wikipedia} was first used to accommodate long documents from a coarse extractive summarizer. Later,~\citet{Zhao2020SEALSE} also focus on long-document summarization and train a joint extractive-abstractive model by weakly supervising the extractor through pseudo labels. This model, although interpretable, does poorly on a dataset like CNN/DM. \cite{pilault-etal-2020-extractive} introduce another explainable summarizing model for long documents by performing a simple extractive step to condition the decoder. They show that
this approach produces more abstractive summaries compared with the copy mechanism.
Unlike these models, we train both modules jointly using the theoretically grounded IB principle with no pseudo labels. Moreover, we seek consistent models suitable for more extractive datasets and achieve results on par with the abstractive module on its own while only using half of the input.
\cite{gehrmann-etal-2018-bottom} trained a content selector separately to tag the words and then use bottom-up attention to only copy words from the tagged set. Similar to our token-level model, this is not useful evidence.

Compressive summarization is another way to have a trade-off between extractive and abstractive methods where extractive summaries are compressed to form the final summary~\cite{mendes-etal-2019-jointly}.   Recently,~\citet{desai-etal-2020-compressive} 
 use syntactic rules to find a high-recall candidate set and then use the notions of plausibility and salience to ensure the grammaticality and importance of the remaining pieces, respectively. Unlike compressive summarization, we explore an extractive-abstractive framework where a concise abstractive summary can be traced back to the evidence; learned jointly with no manual rules or  postprocessing.

\section{Conclusion}
In this paper, we introduced EASE, an extractive-abstractive framework for summarization tasks that trains an extractor and an abstractor in an end-to-end fashion. The extracted evidence forms an explanation of the summary generated by the abstractor. We show that sentence-level extractive-abstractive summarization systems are either on-par or only slightly lower in quality compared to BART, an abstractive model, while at the same time, extracting interpretable explanations for the summary. 
% Future research needs to study the impact of explainability on understanding the source of summaries.

%explored token and sentence level extraction methods from which an abstractive summary is generated. We used soft masking to train the model end-to-end and showed the effectiveness of pretraining for the token-level extraction. We showed that our model generates summaries that have similar quality to  state-of-the-art abstractive summarization systems for more extractive datasets while providing a condensed evidence for its generation.

\bibliographystyle{acl_natbib}
\bibliography{anthology,acl2021}

%\appendix

\end{document}